\documentclass[11pt,a4paper]{article}
\usepackage[hyperref]{emnlp-ijcnlp-2019}
\usepackage{times}
\usepackage{latexsym}
\usepackage{amsmath}
\usepackage{amssymb}
\usepackage{float}
\usepackage{url}
\usepackage{graphicx}
\usepackage{multirow}
\usepackage{booktabs}
\usepackage{bm}
\usepackage{textcomp}
\usepackage{algorithm,algpseudocode}
\graphicspath{ {images/} }
\usepackage{subfig}
\usepackage{xspace}

\aclfinalcopy % Uncomment this line for the final submission

\renewcommand{\textrightarrow}{$\rightarrow$}

\title{Transfer Learning Between Related Tasks \\Using Expected Label Proportions}

\author{Matan Ben Noach$^\ast\dagger$ \and Yoav Goldberg$^\ast\ddagger$ \\
$^\ast$Computer Science Department, Bar-Ilan University, Ramat-Gan Israel\\
$^\dagger$Intel AI Lab, Petah-Tikva Israel\\
$^\ddagger$Allen Institute for Artificial Intelligence\\ \texttt{matan.ben.noach@intel.com, yoav.goldberg@gmail.com} \\
}

\date{}

\begin{document}
\maketitle
\begin{abstract}
Deep learning systems thrive on abundance of labeled training data but such data is not always available, calling for alternative methods of supervision. One such method is expectation regularization (XR) \cite{xr:07}, where models are trained based on expected label proportions. We propose a novel application of the XR framework for transfer learning between related tasks, where knowing the labels of task A provides an estimation of the label proportion of task B. We then use a model trained for A to label a large corpus, and use this corpus with an XR loss to train a model for task B. To make the XR framework applicable to large-scale deep-learning setups, we propose a stochastic batched approximation procedure. We demonstrate the approach on the task of Aspect-based Sentiment classification, where we effectively use a sentence-level sentiment predictor to train accurate aspect-based predictor. The method improves upon fully supervised neural system trained on aspect-level data, and is also cumulative with LM-based pretraining, as we demonstrate by improving a BERT-based Aspect-based Sentiment model.
\end{abstract}

\section{Introduction}
Data annotation is a key bottleneck in many data driven algorithms. Specifically, deep learning models, which became a prominent tool in many data driven tasks in recent years, require large datasets to work well. However, many tasks require manual annotations which are relatively hard to obtain at scale. An attractive alternative is lightly supervised learning \cite{SchapireRRG:02,JinL:05,chang:07,GracaGT:07,QuadriantoSCL:09,MannM:10,GanchevGGT10,Hope2016}, in which the objective function is supplemented by a set of domain-specific soft-constraints over the model's predictions on unlabeled data. For example, in \emph{label regularization} \cite{xr:07}
the model is trained to fit the \emph{true label proportions} of an unlabeled dataset.
Label regularization is special case of \emph{expectation regularization} (XR) \cite{xr:07}, in which the model is trained to fit the conditional probabilities of labels given features.

In this work we consider the case of correlated tasks, in the sense that knowing the labels for task A provides information on the expected label composition of task B. We demonstrate the approach using sentence-level and aspect-level sentiment analysis, which we use as a running example: knowing that a \emph{sentence} has positive sentiment label (task A), we can expect that \emph{most} aspects within this sentence (task B) will also have positive label. While this expectation may be noisy on the individual example level, it holds well in aggregate: given a \emph{set} of positively-labeled sentences, we can robustly estimate the \emph{proportion} of positively-labeled aspects within this set.  For example, in a random set of positive sentences, we expect to find 90\% positive aspects, while in a set of negative sentences, we expect to find 70\% negative aspects. These proportions can be easily either guessed or estimated from a small set.

We propose a novel application of the XR framework for transfer learning in this setup. % tasks, where knowing the labels of task A provides an estimation of expected label proportions for task B.  
% We then use a model trained for A to label a large corpus, and use this corpus with an XR loss to train a model for task B.
We present an algorithm (Sec \ref{sec:method}) that, given a corpus labeled for task A (sentence-level sentiment), learns a classifier for performing task B (aspect-level sentiment) instead, \emph{without} a direct supervision signal for task B. We note that the label information for task A is only used at training time. Furthermore, due to the stochastic nature of the estimation, the task A labels need not be fully accurate, allowing us to make use of \emph{noisy} predictions which are assigned by an automatic classifier (Sections \ref{sec:method} and \ref{sec:app2absc}). In other words, given a medium-sized sentiment corpus with sentence-level labels, and a large collection of \emph{un-annotated} text from the same distribution, we can train an accurate aspect-level sentiment classifier.

The XR loss allows us to use task A labels for training task B predictors. This ability seamlessly integrates into other semi-supervised schemes: we can use the XR loss on top of a pre-trained model to fine-tune the pre-trained representation to the target task, and we can also take the model trained using XR loss and plentiful data and fine-tune it to the target task using the available small-scale annotated data. In Section \ref{sec:pretraining} we explore these options and show that our XR framework improves the results also when applied on top of a pre-trained \textsc{Bert}-based model \cite{bert:18}.

Finally, to make the XR framework applicable to large-scale deep-learning setups, we propose a stochastic batched approximation procedure (Section \ref{sec:sbaxr}). Source code
is available at \url{https://github.com/MatanBN/XRTransfer}.

\section{Background and Related Work}

\subsection{Lightly Supervised Learning}
An effective way to supplement small annotated datasets is to use lightly supervised learning, in which the objective function is supplemented by a set of domain-specific soft-constraints over the model's predictions on unlabeled data. Previous work in lightly-supervised learning focused on training classifiers by using prior knowledge of label proportions \citep{JinL:05, chang:07,musicant:07,xr:07,quadrianto:09,liang:09,GanchevGGT10,mccallum:10,chang:12,Wang:12,zhu:14,Hope2016} or prior knowledge of features label associations \citep{SchapireRRG:02,klien:06,druck:08,melville:09, twit:15}. In the context of NLP,  
%lightly supervised learning was used in a few different setups.
\citet{klien:06} suggested to use distributional similarities of words to train sequence models for part-of-speech tagging and a classified ads information extraction
task. \citet{melville:09} used background lexical information in
terms of word-class associations to train a sentiment classifier. \citet{ganch:13,Wang:14} suggested to exploit the bilingual correlations between a resource rich language and a resource poor language to train a classifier for the resource poor language in a lightly supervised manner.
\subsection{Expectation Regularization (XR)} \label{sec:xr}
\emph{Expectation Regularization} (XR) \cite{xr:07} is a lightly supervised learning method, in which the model is trained to fit the conditional probabilities of labels given features. In the context of NLP, XR was used by \citet{twit:15} to train twitter-user attribute prediction using hundreds of noisy distributional expectations based on census demographics. Here, we suggest using XR to train a target task (aspect-level sentiment) based on the output of a related source-task  classifier (sentence-level sentiment).

\paragraph{Learning Setup}
The main idea of XR is moving from a fully supervised situation in which each data-point $x_i$ has an associated label $y_i$, to a setup in which sets of data points $U_j$ are associated with corresponding label proportions $\mathbf{\tilde{p}_j}$ over that set.

Formally, let $X=\{x_1,x_2,\dots,x_n\} \subseteq{\mathcal{X}}$  be a set of data points, $\mathcal{Y}$ be a set of $|\mathcal{Y}|$ class labels, $U= \{U_1, U_2, \dots, U_m\}$ be a set of sets where $U_j \subseteq X$ for every $j \in \{1,2,\dots,m\}$, and let $\mathbf{\tilde{p}_j} \in R^{|\mathcal{Y}|}$ be the label distribution of set $U_j$. For example, $\mathbf{\tilde{p}_j}=\{.7,.2,.1\}$ would indicate that 70\% of data points in $U_j$ are expected to have class 0, 20\% are expected to have class 1 and 10\% are expected to have class 2. 
Let $p_\theta(x)$ be a parameterized function with parameters $\theta$ from $\mathcal{X}$ to a vector of conditional probabilities over labels in $\mathcal{Y}$. We write $p_\theta(y|x)$ to denote the probability assigned to the $y$th event (the conditional probability of $y$ given $x$).

A typically objective when training on fully labeled data of $(x_i,y_i)$ pairs is to maximize likelihood of labeled data using the cross entropy loss, \[L_{cross}(\theta)=-\sum\limits_i^n\log p_\theta(y_i|x_i)\] % \hat{p}_{it}$ where $t$ is the true label of example $i$ and $\theta$ are the model parameters. 
Instead, in XR our data comes in the form of pairs $(U_j, \mathbf{\tilde{p}_j})$ of sets and their corresponding expected label proportions, and 
we aim to optimize $\theta$ to fit the label distribution $\mathbf{\tilde{p}_j}$ over $U_j$, for all $j$. 
\paragraph{XR Loss} As counting the number of predicted class labels over a set $U$ leads to a non-differentiable objective, \citet{xr:07} suggest to relax it and use instead the model's posterior distribution $\mathbf{\hat{p}_j}$ over the set: 
\begin{equation}
\mathbf{\hat{q}_j}(y)=\sum_{x \in U_j} p_\theta(y|x) 
\label{eq:qj}
\end{equation}
\begin{equation}
\mathbf{\hat{p}_j}(y)=\dfrac{\mathbf{\hat{q}_j}(y)}{\sum_{y'}\mathbf{\hat{q}_j}(y')} 
\end{equation}
where $\mathbf{q}(y)$ indicates the $y$th entry in $\mathbf{q}$.
Then, we would like to set $\theta$ such that  $\mathbf{\hat{p}_j}$ and $\mathbf{\tilde{p}_j}$ are close. \citet{xr:07} suggest to use KL-divergence for this. KL-divergence is composed of two parts:
\[ D_{KL}(\mathbf{\tilde{p}_j}||\mathbf{\hat{p}_j})=-\mathbf{\tilde{p}_j} \cdot \log \mathbf{\hat{p}_j} + \mathbf{\tilde{p}_j} \cdot \log \mathbf{\tilde{p}_j} \] \[= H(\mathbf{\tilde{p}_j},\mathbf{\hat{p}_j}) -H(\mathbf{\tilde{p}_j}) \]
Since $H(\mathbf{\tilde{p}_j})$ is constant, we only need to minimize $H(\mathbf{\tilde{p}_j},\mathbf{\hat{p}_j})$, therefore the loss function becomes:\footnote{Note also that $\forall_{j}|U_j|=1  \iff L_{XR}(\theta)=L_{cross}(\theta)$} 
\begin{equation}
L_{XR}(\theta) = -\sum_{j=1}^m\mathbf{\tilde{p}_j} \cdot \log \mathbf{\hat{p}_j}
\end{equation}

Notice that computing 
$\mathbf{\hat{q}_j}$ requires summation over $p_\theta(x)$ for the entire set $U_j$, which can be prohibitive. We present batched approximation (Section \ref{sec:sbaxr}) to overcome this.

\paragraph{Temperature Parameter}
\citet{xr:07} find that XR might find a degenerate solution. For example, in a three class classification task, where $\tilde{p}_j=\{.5,.35,.15\}$, it might find a solution such that $\hat{p}_\theta(y)=\{.5,.35,.15\}$ for every instance, as a result, every instance will be classified the same. To avoid this, \citet{xr:07} suggest to penalize flat distributions by using a temperature coefficient T likewise:
\begin{equation}
    p_{\theta}(y|x)=\bigg(\frac{e^{\mathbf{z}W+\mathbf{b}}}{\sum\limits_{k}e^{(\mathbf{z}W+\mathbf{b})_k}}\bigg)^{\frac{1}{T}} 
\end{equation}
Where \textbf{z} is a feature vector and W and \textbf{b} are the linear classifier parameters.

\subsection{Aspect-based Sentiment Classification}
In the aspect-based sentiment classification (ABSC) task, we are given a sentence and an aspect, and need to determine the sentiment that is expressed towards the aspect. 
For example the sentence \textit{``Excellent food, although the interior could use some help.``}  has two aspects: \emph{food} and \emph{interior}, a positive sentiment is expressed about the food, but a negative sentiment is expressed about the interior.
A sentence $\alpha=(w_1,w_2,\dots,w_n)$, may contain 0 or more aspects $a_i$, where each aspect corresponds to a sub-sequence of the original sentence, and has an associated sentiment label (\textsc{Neg}, \textsc{Pos}, or \textsc{Neu}). 
Concretely, we follow the task definition in the SemEval-2015 and SemEval-2016 shared tasks \cite{semeval:15,semeval:16}, in which the relevant aspects are given and the task focuses on finding the sentiment label of the aspects.  

While sentence-level sentiment labels are relatively easy to obtain, aspect-level annotation are much more scarce, as demonstrated in the small datasets of the SemEval shared tasks.

\section{Technical Contributions}
\subsection{Transfer-training between related tasks with XR}\label{sec:method}

\begin{algorithm}[t!] 
{\bf Inputs}: A dataset $(U_1,...,U_m,\mathbf{\tilde{p}_1},...,\mathbf{\tilde{p}_m})$, batch size $k$, differentiable classifier $p_\theta(y|x)$
\begin{algorithmic}[H]
\While{not converged}
\State $j \gets$ random($1,...,m$)
\State $U' \gets$ random-choice($U_j$,$k$)
\State $\mathbf{\hat{q}_u'
}\gets \sum_{x\in U'}p_\theta(x)$
%\State $\hat{p}_u \gets$ $\frac{\hat{q}_u}{\sum_{y}\hat{q}_u(y)}$
\State $\mathbf{\hat{p}_u'} \gets$ $\text{normalize}(\mathbf{\hat{q}_u'})$
\State $\ell \gets -\tilde{p}_j\log\hat{p}_u$ 
\Comment{Compute loss $\ell$ (eq (4))}
\State Compute gradients and update $\theta$
\EndWhile
\State \Return{$\theta$}
\end{algorithmic}
\caption{Stochastic Batched XR}
\label{alg:sbxr}
\end{algorithm}

Consider two classification tasks over a shared input space, a source task $s$ from $\mathcal{X}$ to $\mathcal{Y}^s$ and a target task $t$ from $\mathcal{X}$ to $\mathcal{Y}^t$, which are related through a conditional distribution $P(y^t=i|y^s=j)$. In other words, a labeling decision for task $s$ induces an expected label distribution over the task $t$. For a set of datapoints $x_1,...,x_n$ that share a source label $y^s$, we expect to see a target label distribution of $P(y^t|y^s)=\mathbf{\tilde{p}_{y^s}}$.

Given a large unlabeled dataset $D^u=(x^u_1,...,x^u_{|D^u|})$, a small labeled dataset for the target task $D^t = ((x^t_1,y^t_1),...,(x^t_{|D^t|},y^t_{|D^t|}))$, classifier $C^s: \mathcal{X} \mapsto \mathcal{Y}^s$ (or sufficient training data to train one) for the source task,\footnote{Note that the classifier does not need to be trainable or differentiable. It can be a human, a rule based system, a non-parametric model, a probabilistic model, a deep learning network, etc. In this work, we use a neural classification model.} we wish to use $C^s$ and $D^u$ to train a good classifier $C^t:\mathcal{X}\mapsto \mathcal{Y}^t$ for the target task.
This can be achieved using the following procedure.

\begin{itemize}
\item Apply $C^s$ to $D^t$, resulting in a noisy source-side labels $\tilde{y}^s_i = C^s(x^t_i)$ for the target task.
\item Estimate the conditional probability $P(y^t|\tilde{y}^s)$ table using MLE estimates over $D^t$ 
\[\tilde{p}_j(y^t=i|\tilde{y}^s=j) = \frac{\#(y^t=i, \tilde{y}^s=j)}{\#(\tilde{y}^s=j)}\] 
where $\#$ is a counting function over $D^t$.\footnote{In theory, we could estimate---or even ``guess''---these $|\mathcal{Y}^s|\times|\mathcal{Y}^t|$ values without using $D^t$ at all. In practice, and in particular because we care about the target label proportions given \emph{noisy} source labels $\tilde{y}^s$ assigned by $C^s$, we use MLE estimates over the tagged $D^t$.}
\item Apply $C^s$ to the unlabeled data $D^u$ resulting in labels $C^s(x^u_i)$. Split $D^u$ into $|\mathcal{Y}^s|$ sets $U_j$ according to the labeling induced by $C^s$:
\[
U_j = \{x^u_i \mid x^u_i \in D^u \wedge C^s(x^u_i)=j\}
\]
\item Use Algorithm \ref{alg:sbxr} to train a classifier for the target task using input pairs $(U_j,\mathbf{\tilde{p}_j})$ and the XR loss.
\end{itemize}

In words, by using XR training, we use the expected label proportions over the target task given predicted labels of the source task, to train a target-class classifier.

\subsection{Stochastic Batched Training for Deep XR}\label{sec:sbaxr}
\citet{xr:07} and following work take the base classifier $p_\theta(y|x)$ to be a logistic regression classifier, for which they manually derive gradients for the XR loss and train with LBFGs \cite{lbfgs:95}. However, nothing precludes us from using an arbitrary neural network instead, as long as it culminates in a softmax layer.

One complicating factor is that the computation of $\mathbf{\hat{q}_j}$ in equation (\ref{eq:qj}) requires a summation over $p_\theta(x)$ for the entire set $U_j$, which in our setup may contain hundreds of thousands of examples, making gradient computation and optimization impractical. We instead proposed a \emph{stochastic batched approximation} in which, instead of requiring that the full constraint set $U_j$ will match the expected label posterior distribution, we require that sufficiently large random subsets of it will match the distribution. At each training step we compute the loss and update the gradient with respect to a different random subset. Specifically, 
% in which we approximate the full gradient over $U_j$ using random subsets of $U_j$. Specifically, 
in each training step we sample a random pair $(U_j,\mathbf{\tilde{p}_j})$, sample a random subset $U'$ of $U_j$ of size $k$, and compute the local XR loss of set $U'$:
\begin{equation}
    L_{XR}(\theta;j,U')=
    %-\sum_y \tilde{p}_j(y)\log\hat{p}_{u'}(y)=
    -\mathbf{\tilde{p}_j} \cdot \log \mathbf{\hat{p}_{u'}} \end{equation} 
    where $\mathbf{\hat{p}_{u'}}$ is computed by summing over the elements of $U'$ rather than of $U_j$ in equations (\ref{eq:qj}--2).
The stochastic batched XR training algorithm is given in Algorithm \ref{alg:sbxr}. For large enough $k$, the expected label distribution of the subset is the same as that of the complete set.

\begin{figure*}[t!]
  \hfill\includegraphics[scale=0.4]{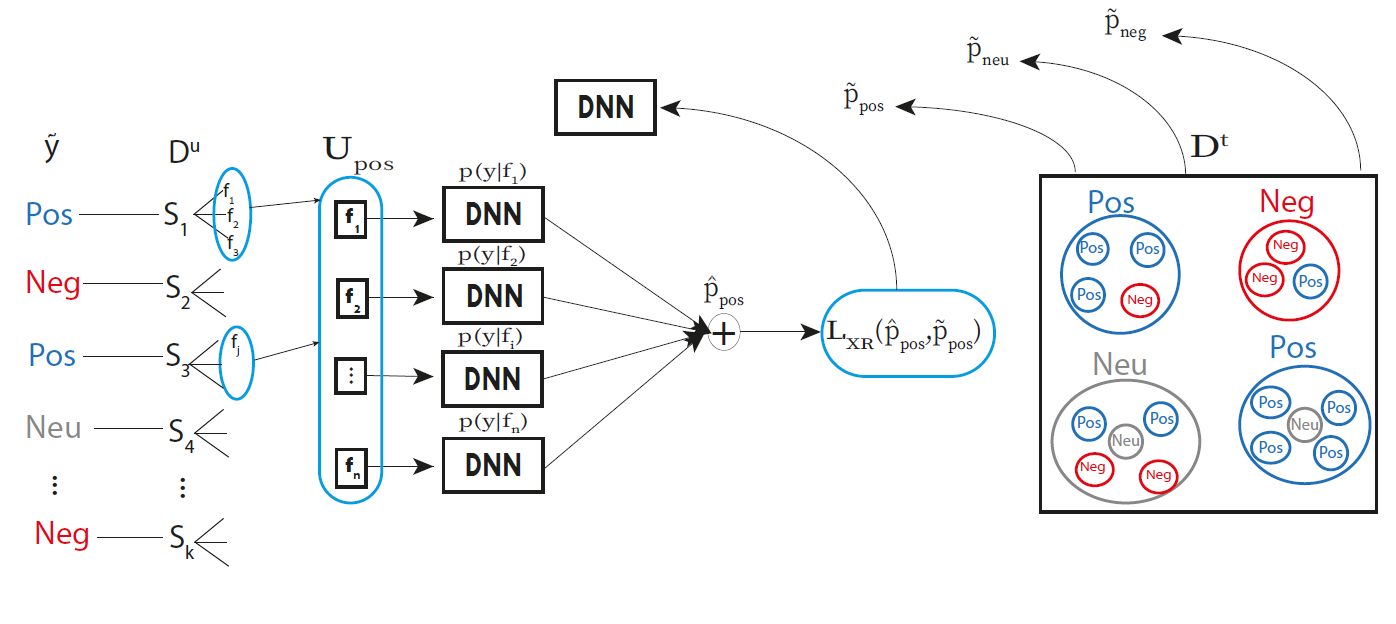}\hspace*{\fill}
  \caption{Illustration of the algorithm. $C^s$ is applied to $D^u$ resulting in $\tilde{y}$ for each sentence, $U_j$ is built according with the fragments of the same labelled sentences, the probabilities for each fragment in $U_j$ are summed and normalized, the XR loss in equation (4) is calculated and the network is updated.
  }
  \label{algo}
\end{figure*}

\begin{figure}[t!]
 \hfill\includegraphics[scale=0.45]{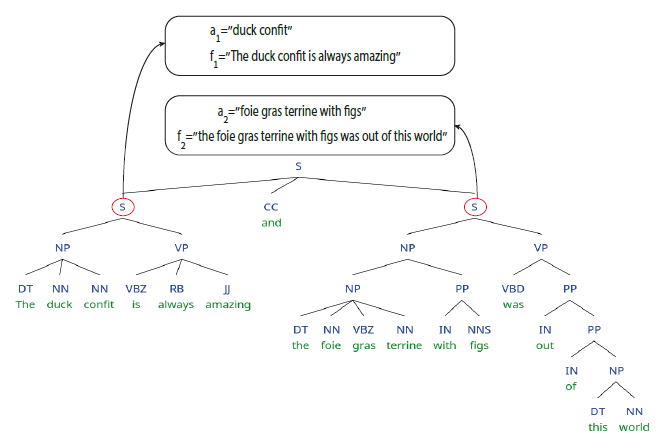}\hspace*{\fill}
 \caption{Illustration of the decomposition procedure, when given $a_1$=``duck confit`` and $a_2$= ``foie gras terrine with figs`` as the pivot phrases.}
 \label{decomp}
\end{figure}

\section{Application to Aspect-based Sentiment}\label{sec:app2absc}
We demonstrate the procedure given above by training Aspect-based Sentiment Classifier (ABSC) using sentence-level\footnote{In practice, our ``sentences'' are in fact short documents, some of which are composed of two or more sentences.} sentiment signals.

\subsection{Relating the classification tasks}
We observe that while the sentence-level sentiment does not \emph{determine} the sentiment of individual aspects (a positive sentence may contain negative remarks about some aspects), it is very predictive of the \emph{proportion} of sentiment labels of the fragments within a sentence.
Positively labeled sentences are likely to have more positive aspects and fewer negative ones, and vice-versa for negatively-labeled sentences. While these proportions may vary on the individual sentence level, we expect them to be stable when aggregating fragments from several sentences: when considering a large enough sample of fragments that all come from positively labeled sentences, we expect the different samples to have roughly similar label proportions to each other. This situation is idealy suited for performing XR training, as described in section \ref{sec:method}.

The application to ABSC is almost straightforward, but is complicated a bit by the decomposition of sentences into fragments: each sentence level decision now corresponds to multiple fragment-level decisions.
Thus, we apply the sentence-level (task A) classifier $C^s$ on the aspect-level corpus $D^t$ by applying it on the sentence level and then associating the predicted sentence labels with each of the fragments, resulting in fragment-level labeling.
Similarly, when we apply $C^s$ to the unlabeled data $D^u$ we again do it at the sentence level, but the sets $U_j$ are composed of fragments, not sentences: \[U_j=\{f^\alpha_i \mid \alpha \in D^u \wedge f^\alpha_i \in \text{frags}(\alpha) \wedge C^s(\alpha)=j\}\]

We then apply algorithm \ref{alg:sbxr} as is: at each step of training we sample a source label $j\in\{$\textsc{Pos,Neg,Neu}$\}$, sample $k$ fragments from $U_j$, and use the XR loss to fit the expected fragment-label proportions over these $k$ fragments to $\mathbf{\tilde{p}_j}$.
Figure \ref{algo} illustrates the procedure.

\subsection{Classification Architecture}
We model the ABSC problem by associating each (sentence,aspect) pair with a \emph{sentence-fragment}, and constructing a neural classifier from fragments to sentiment labels.
We heuristically decompose a sentence into fragments. We use the same BiLSTM based neural architecture for both sentence classification and fragment classification. 

\paragraph{Fragment-decomposition}
\label{sec:decomp}
We now describe the procedure we use to associate a sentence fragment with each (sentence,aspect) pairs.
The shared tasks data associates each aspect with a pivot-phrase $a$, 
where  pivot phrase $(w_1,w_2,...w_l)$ is defined as a pre-determined sequence of words that is contained within the sentence.
For a sentence $\alpha$, a set of pivot phrases $A=(a_1,...,a_m)$ and a specific pivot phrase $a_i$, we consult the constituency parse tree of $\alpha$ and look for tree nodes that satisfy the following conditions:\footnote{Condition (2) coupled with selecting the highest node pushes towards complete phrases that contain opinions (which are usually expressed with adjectives or verbs), while the other conditions focus the attention on the desired pivot phrase.}

\begin{enumerate}
  \item The node governs the desired pivot phrase $a_i$.
  \item The node governs either a verb (VB, VBD, VBN, VBG, VBP, VBZ) or an adjective (JJ, JJR, JJS), which is different than any $a_j \in A$.
  \item The node governs a minimal number of pivot phrases from $(a_1,...,a_m)$, ideally only $a_i$.
\end{enumerate}

We then select the highest node in the tree that satisfies all conditions. The span governed by this node is taken as the fragment associated with aspect $a_i$.\footnote{On the rare occasions where we cannot find such a node, we take the root node of the tree (the entire sentence) as the fragment for the given aspect.}
The decomposition procedure is demonstrated in Figure \ref{decomp}.

When aspect-level information is given, we take the pivot-phrases to be the requested aspects. When aspect-level information is \emph{not available}, we take each noun in the sentence to be a pivot-phrase.

\paragraph{Neural Classifier} 
\label{sec:bilstm}
Our classification model is a simple 1-layer BiLSTM encoder (a concatenation of the last states of a forward and a backward running LSTMs) followed by a linear-predictor.
The encoder is fed either a complete sentence or a sentence fragment. 

\section{Experiments}

\paragraph{Data} Our \emph{target} task is aspect-based fragment-classification, with small labeled datasets from the SemEval 2015 and 2016 shared tasks, each dataset containing aspect-level predictions for about 2000 sentences in the restaurants reviews domain. Our \emph{source} classifier is based on training on up to 10,000 sentences from the same domain and 2000 sentences for validation, labeled for only for sentence-level sentiment. We additionally have an unlabeled dataset of up to 670,000 sentences from the same domain\footnote{All of the sentence-level sentiment data is obtained from the Yelp dataset challenge: \url{https://www.yelp.com/dataset/challenge}}. We tokenized all datasets using the Tweet Tokenizer from NLTK package\footnote{\url{https://www.nltk.org/}} and parsed the tokenized sentences with AllenNLP parser.\footnote{\url{https://allennlp.org/}}

\paragraph{Training Details}
Both the sentence level classification models and the models trained with XR have a hidden state vector dimension of size 300, they use dropout \cite {dropout:12} on the sentence representation or fragment representation vector (rate=0.5) and optimized using Adam \cite{adam:14}. The sentence classification is trained  with a batch size of 30 and XR models are trained with batch sizes $k$ that each contain 450 fragments\footnote{We also increased the batch sizes of the baselines to match those of the XR setups. This decreased the performance of the baselines, which is consistent with the folk knowledge in the community according to which smaller batch sizes are more effective overall.}. We used a temperature parameter of 1\footnote{Despite \cite{xr:07} claim regarding the temperature parameter, we observed lower performance when using it in our setup. However, in other setups this parameter might be found to be beneficial.}.
We use pre-trained 300-dimensional GloVe embeddings\footnote{\url{https://nlp.stanford.edu/projects/glove/}} \cite{glove:14}, and fine-tune them during training. 
The XR training was validated with a validation set of 20\% of SemEval-2015 training set, the sentence level BiLSTM classifiers were validated with a validation of 2000 sentences.\footnote{We also tested the sentence BiLSTM baselines with a SemEval validation set, and received slightly lower results without a significant statistical difference.} When fine-tuning to the aspect based task we used 20\% of train in each dataset as validation and evaluated on this set. On each training method the models were evaluated on the validation set, after each epoch and the best model was chosen. The data is highly imbalanced, with only very few sentences receiving a \textsc{Neu} label. We do not deal with this imbalance directly and train both the sentence level and the XR aspect-based training on the imbalanced data. However, when training $C^s$, we trained five models and chose the best model that predicts correctly at least 20\% of the neutral sentences.
The models are implemented using DyNet\footnote{\url{https://github.com/clab/dynet}} \cite{dynet:17}.  

\paragraph{Baseline models}
\begin{table*}[t!]
\centering
\small
\scalebox{0.88}{
\begin{tabular}{llccccc}
\toprule 
\multirow{ 2}{*}{Data} &\multirow{ 2}{*}{Method} & \multicolumn{2}{c}{SemEval-15} & \multicolumn{2}{c}{SemEval-16} & \\\cline{3-6}
&&Acc. &Macro-F1 &Acc. &Macro-F1\\\hline
A&TDLSTM+ATT \cite{td-lstm:16}  & 77.10 & 59.46 & 83.11 & 57.53 \\
A&ATAE-LSTM \cite{atae:16} &78.48 &62.84 &83.77 &61.71 \\
A&MM \cite{aspmem:16} &77.89 &59.52 &83.04 &57.91 \\
A&RAM \cite{att-mem:17} &79.98 &60.57 &83.88 &62.14\\
A&LSTM+SynATT+TarRep \cite{eff-att:18} & 81.67 & 66.05 & 84.61 & 67.45\\
\hline
S+A&Semisupervised \cite{exploit:18} &  81.30 & \textbf{68.74} & 85.58 & 69.76 \\
S&BiLSTM-$10^4$ Sentence Training & 80.24 $\pm$ 1.64 & 61.89 $\pm$ 0.94& 80.89 $\pm$ 2.79& 61.40 $\pm$ 2.49\\
S+A&BiLSTM-$10^4$ Sentence Training \textrightarrow Aspect Based Finetuning  & 77.75 $\pm$ 2.09 & 60.83 $\pm$ 4.53& 84.87$\pm$ 0.31& 61.87 $\pm$ 5.44\\
\hline
N&BiLSTM-XR-Dev Estimation & \textbf{83.31$^*\pm$ 0.62} &  62.24 $\pm$ 0.66 & \textbf{87.68$^*\pm$ 0.47} & 63.23 $\pm$ 1.81\\
N&BiLSTM-XR & \textbf{83.31$^*\pm$ 0.77} & 64.42 $\pm$ 2.78 & \textbf{88.12$^*\pm$ 0.24} & 68.60 $\pm$ 1.79\\
N+A&BiLSTM-XR \textrightarrow Aspect Based Finetuning & \textbf{83.44$^*\pm$ 0.74} & \textbf{67.23 $\pm$ 1.42} & \textbf{87.66$^*\pm$ 0.28} & \textbf{71.19\textdagger $\pm$ 1.40}\\
\bottomrule

\end{tabular}
}
\caption{Average accuracies and Macro-F1 scores over five runs with random initialization along with their standard deviations. Bold: best results or within std of them. $^*$ indicates that the method's result is significantly better than all baseline methods, \textbf{\textdagger} \xspace indicates that the method's result is significantly better than all baselines methods that use the aspect-based data only, with $p < 0.05$ according to a one-tailed unpaired t-test. 
The data annotations \textbf{S}, \textbf{N} and \textbf{A} indicate training with Sentence-level, Noisy sentence-level and Aspect-level data respectively.
Numbers for TDLSTM+Att,ATAE-LSTM,MM,RAM and LSTM+SynATT+TarRep are from \cite{eff-att:18}. Numbers for Semisupervised are from \cite{exploit:18}.}\label{tbl:compare}
\end{table*}

In recent years many neural network architectures with increasing sophistication were applied to the ABSC task \cite{phrasernn:15,tsc:15,td-lstm:16,aspmem:16,atae:16,gated:16,hier:16,review-terror:17,amts:17,att-mem:17,rec-ent:18,feature-en:18,target-s:18,clauses:18,convnn:18,multi-g:18,hier-att:18,depandatt:18}.
We compare to a series of state-of-the-art ABSC neural classifiers that participated in the shared tasks. TDLSTM-ATT \cite{td-lstm:16} encodes the information around an aspect using forward and backward LSTMs, followed by an attention mechanism. 
ATAE-LSTM \cite{atae:16} is an attention based LSTM variant. MM \cite{aspmem:16} is a deep memory network with multiple-hops of attention layers. RAM \cite{att-mem:17} uses multiple attention mechanisms combined with a recurrent neural networks and a weighted memory mechanism. LSTM+SynATT+TarRep \cite{eff-att:18} is an attention based LSTM which incorporates syntactic information into the attention mechanism and uses an auto-encoder structure to produce an aspect representations. All of these models are trained only on the small, fully-supervised ABSC datasets. 

``Semisupervised'' is the semi-supervised setup of \cite{exploit:18}, it trains an attention-based LSTM model on 30,000 documents additional to an aspect-based train set, 10,000 documents to each class.
We consider additional two simple but strong semi-supervised baselines. Sentence-BiLSTM is our BiLSTM model trained on the $10^4$ sentence-level annotations, and applied as-is to the individual fragments. Sentence-BiLSTM+Finetuning is the same model, but fine-tuned on the aspect-based data as explained above. Finetuning is performed using our own implementation of the attention-based model of \citet{exploit:18}.\footnote{We changed the LSTM component to a BiLSTM.} Both these models are on par with the fully-supervised ABSC models.
\paragraph{Empirical Proportions}\label{prop}
The proportion constraint sets $\mathbf{\tilde{p}_j}$ based on the SemEval-2015 aspect-based train data are:\\
$\mathbf{\tilde{p}_\textsc{pos}}=\{\textsc{Pos}: 0.93, \textsc{Neg}: 0.06, \textsc{Neu}: 0.01\}$\\
$\mathbf{\tilde{p}_\textsc{neg}}=\{\textsc{Pos}: 0.27, \textsc{Neg}: 0.7, \textsc{Neu}: 0.03\}$\\
$\mathbf{\tilde{p}_\textsc{neu}}=\{\textsc{Pos}: 0.45, \textsc{Neg}: 0.41, \textsc{Neu}: 0.14\}$\\

\subsection{Main Results}\label{sec:results}

Table \ref{tbl:compare} compares these baselines to three XR conditions.\footnote{To be consistent with existing research \cite{exploit:18}, aspects with conflicted polarity are removed.}

The first condition, BiLSTM-XR-Dev, performs XR training on the automatically-labeled sentence-level dataset. The only access it has to aspect-level annotation is for estimating the proportions of labels for each sentence-level label, which is done based on the validation set of SemEval-2015 (i.e., 20\% of the train set). The XR setting is very effective:
without using any in-task data, this model already surpasses all other models, both supervised and semi-supervised, except for the \cite{exploit:18,eff-att:18} models which achieve higher F1 scores. We note that in contrast to XR, the competing models have complete access to the supervised aspect-based labels.  
The second condition, BiLSTM-XR, is similar but now the model is allowed to estimate the conditional label proportions based on the entire aspect-based training set (the classifier still does not have direct access to the labels beyond the aggregate proportion information).  This improves results further, showing the importance of accurately estimating the proportions. Finally, in BiLSTM-XR+Finetuning, we follow the XR training with fully supervised fine-tuning on the small labeled dataset, using the attention-based model of \citet{exploit:18}. This achieves the best results, and surpasses also the semi-supervised \citet{exploit:18} baseline on accuracy, and matching it on F1.\footnote{We note that their setup uses clean and more balanced annotations, i.e. they use 10,000 samples for each label, which helps predicting the infrequent neutral sentiment. We however, use noisy sentence sentiment labels which are automatically obtained from a trained classifier, which trains on 10,000 samples in their natural imbalanced distribution.}

We report significance tests for the robustness of the method under random parameter initialization.
Our reported numbers are averaged over five random initialization. Since the datasets are unbalanced w.r.t the label distribution, we report both accuracy and macro-F1.

The XR training is also more stable than the other semi-supervised baselines, achieving substantially lower standard deviations across different runs.

\subsection{Further experiments}\label{sec:analysis}
\begin{figure*}[!tbp]
\centering
\begin{tabular}{cccc}

\hfill\includegraphics[width=0.22\textwidth]{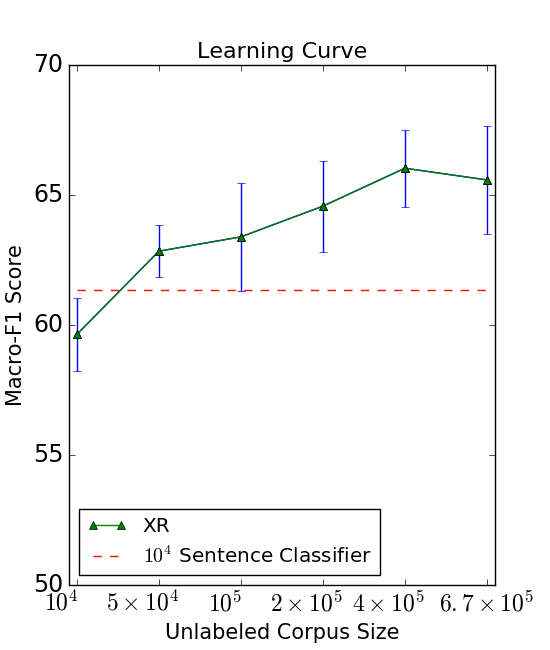} &
\includegraphics[width=0.22\textwidth]{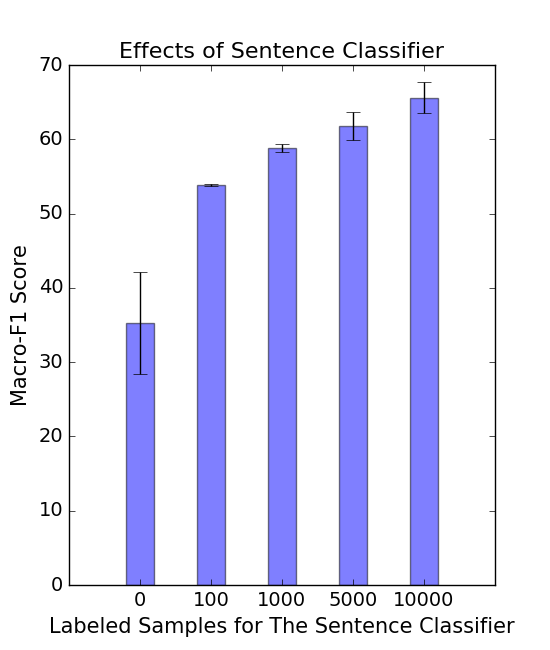} &
\includegraphics[width=0.22\textwidth]{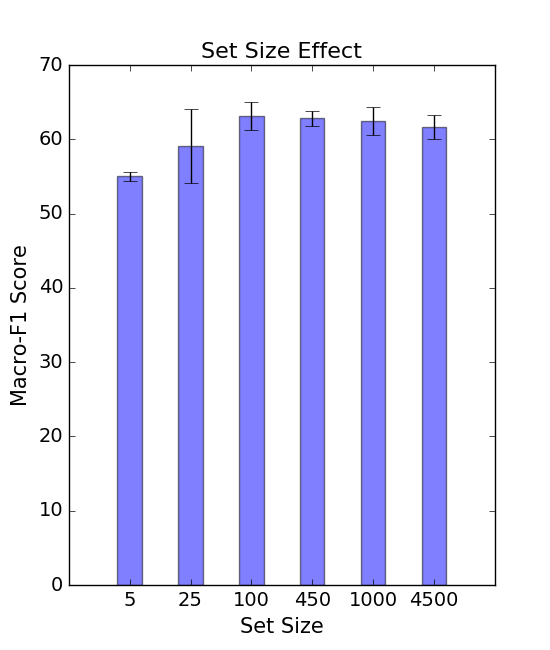} &  
\includegraphics[width=0.22\textwidth]{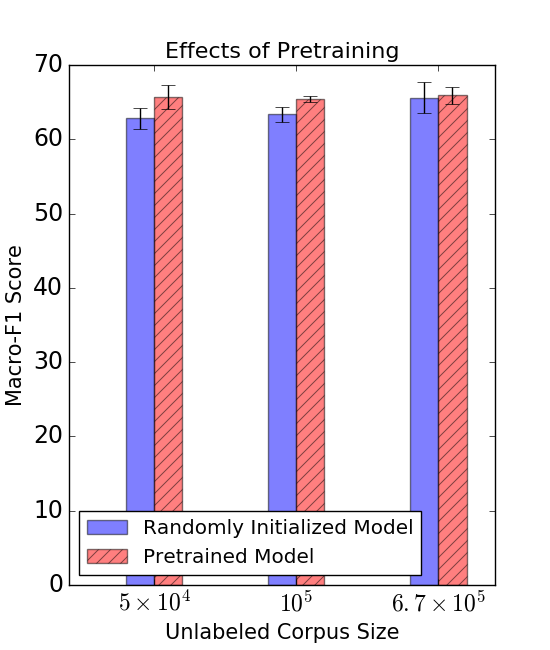} \\
(a)  & (b) & (c) & (d) 
\end{tabular} 
\caption{Macro-F1 scores for the entire SemEval-2016 dataset of the different analyses. (a) the contribution of unlabeled data. (b) the effect of sentence classifier quality. (c) the effect of k. (d) the effect of sentence-level pretraining vs. corpus size.}
\label{experiments}
\end{figure*}

In each experiment in this section we estimate the proportions using the SemEval-2015 train set.
\paragraph{Effect of unlabeled data size}
How does the XR training scale with the amount of unlabeled data? Figure \ref{experiments}a shows the macro-F1 scores on the entire SemEval-2016 dataset, with different unlabeled corpus sizes (measured in number of sentences).
An unannotated corpus of $5 \times 10^4$ sentences is sufficient to surpass the results of the $10^4$ sentence-level trained classifier, and more unannotated data further improves  the results.

\paragraph{Effect of Base-classifier Quality}
Our method requires a sentence level classifier $C^s$ to label both the target-task corpus and the unlabeled corpus. How does the quality of this classifier affect the overall XR training? We vary the amount of supervision used to train $C^s$ from 0 sentences (assigning the same label to all sentences), to 100, 1000, 5000 and 10000 sentences. We again measure macro-F1 on the entire SemEval 2016 corpus.

The results in Figure \ref{experiments}b show that when using the prior distributions of aspects (0), the model struggles to learn from this signal, it learns mostly to predict the majority class, and hence reaches very low F1 scores of 35.28. The more data given to the sentence level classifier, the better the potential results will be when training with our method using the classifier labels, with a classifiers trained on 100,1000,5000 and 10000 labeled sentences, we get a F1 scores of 53.81, 58.84, 61.81, 65.58 respectively. Improvements in the source task classifier's quality clearly contribute to the target task accuracy.

\paragraph{Effect of $k$}
The Stochastic Batched XR algorithm (Algorithm \ref{alg:sbxr}) samples a batch of $k$ examples at each step to estimate the posterior label distribution used in the loss computation. How does the size of $k$ affect the results? We use $k=450$ fragments in our main experiments, but smaller values of $k$ 
reduce GPU memory load and may train better in practice.
We tested our method with varying values of $k$ on a sample of $5 \times 10^4$, using batches that are composed of fragments of 5, 25, 100, 450, 1000 and 4500 sentences. The results are shown in Figure \ref{experiments}c. Setting $k=5$ result in low scores. Setting $k=25$ yields better F1 score but with high variance across runs. For $k=100$ fragments the results begin to stabilize, we also see a slight decrease in F1-scores with larger batch sizes. We attribute this drop despite having better estimation of the gradients to the general trend of larger batch sizes being harder to train with stochastic gradient methods.

\begin{table*}[t!]
\centering
\small
\scalebox{0.88}{
\begin{tabular}{lcccccc}

\toprule 
\multirow{ 2}{*}{Data}&\multirow{ 2}{*}{Training} & \multicolumn{2}{c}{SemEval-15} & \multicolumn{2}{c}{SemEval-16} & \\\cline{3-6}
&&Acc. &Macro-F1 &Acc. &Macro-F1\\\hline\hline
N&BiLSTM-XR & 83.31 $\pm$ 0.77 & 64.42 $\pm$ 2.78 & 88.12 $\pm$ 0.24 & 68.60 $\pm$ 1.79\\
N+A&BiLSTM-XR \textrightarrow Aspect Based Finetuning & 83.44 $\pm$ 0.74 & 67.23 $\pm$ 1.42 & 87.66 $\pm$ 0.28 & 71.19 $\pm$ 1.40\\
\hline
A&\textsc{Bert}\textrightarrow  Aspect Based Finetuning & 81.87 $\pm$ 1.12 & 59.24 $\pm$ 4.94 & 85.81 $\pm$ 1.07 & 62.46 $\pm$ 6.76 \\
S&\textsc{Bert}\textrightarrow$10^4$ Sent Finetuning & 83.29 $\pm$ 0.77 & 66.79 $\pm$ 1.99 & 84.53 $\pm$ 1.66 & 65.53 $\pm$ 3.03\\
S+A&\textsc{Bert}\textrightarrow  $10^4$ Sent Finetuning \textrightarrow  Aspect Based Finetuning & 82.54 $\pm$ 1.21 & 64.13 $\pm$ 5.05 & 85.67 $\pm$ 1.14 & 64.13 $\pm$ 7.07 \\

\hline
N&\textsc{Bert}\textrightarrow XR & 85.46$^*$ $\pm$ 0.59 & 66.86 $\pm$ 2.8 & 89.5$^*$ $\pm$ 0.55 & 70.86\textdagger $\pm$ 2.96 \\
N+A&\textsc{Bert}\textrightarrow  XR \textrightarrow  Aspect Based Finetuning & \textbf{85.78$^*$ $\pm$ 0.65} & \textbf{68.74 $\pm$ 1.36} & \textbf{89.57$^*$ $\pm$ 1.4} & \textbf{73.89$^*$ $\pm$ 2.05} \\
\bottomrule
\end{tabular}
}
\caption {\textsc{Bert} pre-training: average accuracies and Macro-F1 scores from five runs and their stdev. $\mathbf{^*}$ indicates that the method's result is significantly better than all baseline methods, \textbf{\textdagger} \xspace  indicates that the method's result is significantly better than all non XR baseline methods, with $p < 0.05$ according to a one-tailed unpaired t-test. The data annotations \textbf{S}, \textbf{N} and \textbf{A} indicate   training with Sentence-level, Noisy sentence-level and Aspect-level data respectively.}\label{bert}
\end{table*}

\subsection{Pre-training, \textsc{Bert}}\label{sec:pretraining}
The XR training can be performed also over pre-trained representations. We experiment with two pre-training methods: (1) pre-training by training the BiLSTM model to predict the noisy sentence-level predictions. (2) Using the pre-trained \textsc{Bert} representation \cite{bert:18}.
For (1), we compare the effect of pre-train on unlabeled corpora of sizes of 
$5 \times 10^4$, $10^5$ and $6.7 \times 10^5$ sentences. Results in Figure \ref{experiments}d show that this form of pre-training is effective for smaller unlabeled corpora but evens out for larger ones.

\paragraph{\textsc{Bert}} For the \textsc{Bert} experiments, we experiment with the \textsc{Bert}-base model\footnote{We could not fit $k=450$ sets of \textsc{Bert}-large on our GPU.} with $k=450$ sets, 30 epochs for XR training or sentence level fine-tuning\footnote{When fine-tuning to the sentence level task, we provide the sentence as input. When fine-tuning to the aspect-level task, we provide the sentence, a seperator and then the aspect.} and 15 epochs for aspect based fine-tuning, on each training method we evaluated the model on the dev set after each epoch and the best model was chosen\footnote{The other configuration parameters were the default ones in \url{https://github.com/huggingface/pytorch-pretrained-BERT}}.
We compare the following setups:\\
-\textsc{Bert}\textrightarrow Aspect Based Finetuning: pretrained \textsc{bert} model finetuned to the aspect based task.\\
-\textsc{Bert}$\rightarrow10^4$: A pretrained \textsc{bert} model finetuned to the sentence level task on the $10^4$ sentences, and tested by predicting fragment-level sentiment.\\
-\textsc{Bert}\textrightarrow$10^4$\textrightarrow Aspect Based Finetuning: pretrained \textsc{bert} model finetuned to the sentence level task, and finetuned again to the aspect based one.\\
-\textsc{Bert}\textrightarrow XR: pretrained \textsc{bert} model followed by XR training using our method.\\
-\textsc{Bert}\textrightarrow~XR~\textrightarrow~Aspect Based Finetuning: pretrained \textsc{bert} followed by XR training and then fine-tuned to the aspect level task.

The results are presented in Table \ref{bert}. As before, aspect-based fine-tuning is beneficial for both SemEval-16 and SemEval-15. Training a BiLSTM with XR surpasses pre-trained \textsc{bert} models and using XR training on top of the pre-trained \textsc{Bert} models substantially increases the results even further.

\section{Discussion}
We presented a transfer learning method based on expectation regularization (XR), and demonstrated its effectiveness for training aspect-based sentiment classifiers using sentence-level supervision. The method achieves state-of-the-art results for the task, and is also effective for improving on top of a strong pre-trained \textsc{Bert} model.
The proposed method provides an additional data-efficient tool in the modeling arsenal,  which can be applied on its own or together with another training method, in situations where there is a conditional relations between the labels of a source task for which we have supervision, and a target task for which we don't.

While we demonstrated the approach on the sentiment domain, the required conditional dependence between task labels is present in many situations. Other possible application of the method includes training language identification of tweets given geo-location supervision (knowing the geographical region gives a prior on languages spoken), training predictors for renal failure from textual medical records given classifier for diabetes (there is a strong correlation between the two conditions), training a political affiliation classifier from social media tweets based on age-group classifiers, zip-code information, or social-status classifiers (there are known correlations between all of these to political affiliation), training hate-speech detection based on emotion detection, and so on. 

\section*{Acknowledgements} The work was supported in part by The Israeli Science Foundation (grant number 1555/15).

\bibliography{main}
\bibliographystyle{acl_natbib}

\end{document}